\documentclass[10pt, a4paper]{article}

\usepackage{lrec-coling2024} 

\usepackage{natbib}
\usepackage{multibib}
\makeatletter
\def\@mb@citenamelist{cite,citep,citet,citealp,citealt,citepalias,citetalias}
\makeatother
\newcites{languageresource}{~}

\usepackage{amsmath,amssymb,amsfonts}
\usepackage{graphicx}
\usepackage{tabularx}
\usepackage{soul}
\usepackage{bbding}
\usepackage{titlesec}
\usepackage{multirow}
\usepackage{booktabs}
\usepackage{colortbl} 
\usepackage{algorithm,algorithmic} 
\titleformat{\section}{\normalfont\large\bfseries\center}{\thesection.}{1em}{}
\titleformat{\subsection}{\normalfont\SmallTitleFont\bfseries\raggedright}{\thesubsection.}{1em}{}
\titleformat{\subsubsection}{\normalfont\normalsize\bfseries\raggedright}{\thesubsubsection.}{1em}{}
\renewcommand\thesection{\arabic{section}}
\renewcommand\thesubsection{\thesection.\arabic{subsection}}
\renewcommand\thesubsubsection{\thesubsection.\arabic{subsubsection}}

\usepackage{xcolor}
\usepackage{hyperref}
 \definecolor{darkblue}{rgb}{0, 0, 0.5}
  \hypersetup{colorlinks=true, citecolor=darkblue, linkcolor=darkblue, urlcolor=darkblue}

\usepackage{xstring}

\usepackage{color}

\title{Towards Human-Like Machine Comprehension: \\ Few-Shot Relational Learning in Visually-Rich Documents}

\name{Hao Wang$^{1}$, Tang Li$^{1}$, Chenhui Chu$^{2}$, Nengjun Zhu$^{1}$ , Rui Wang$^{3,*}$\thanks{* Corresponding author.} and Pinpin Zhu$^{1}$ } 

\address{$^{1}$ School of Computer Engineering and Science, Shanghai University, China \\
 $^{2}$Graduate School of Informatics, Kyoto University, Japan\\
$^{3}$Department of Computer Science and Engineering, Shanghai Jiao Tong University, China\\
          \{wang-hao, ltttlttt, zhu\_nj, zhupp\}@shu.edu.cn, chu@i.kyoto-u.ac.jp,  wangrui12@sjtu.edu.cn\\}

\abstract{Key-value relations are prevalent in Visually-Rich Documents (VRDs), often depicted in distinct spatial regions accompanied by specific color and font styles. These non-textual cues serve as important indicators that greatly enhance human comprehension and acquisition of such relation triplets. However, current document AI approaches often fail to consider this valuable prior information related to visual and spatial features, resulting in suboptimal performance, particularly when dealing with limited examples. To address this limitation, our research focuses on few-shot relational learning, specifically targeting the extraction of key-value relation triplets in VRDs. Given the absence of a suitable dataset for this task, we introduce two new few-shot benchmarks built upon existing supervised benchmark datasets. Furthermore, we propose a variational approach that incorporates relational 2D-spatial priors and prototypical rectification techniques. This approach aims to generate relation representations that are more aware of the spatial context and unseen relation in a manner similar to human perception. Experimental results demonstrate the effectiveness of our proposed method by showcasing its ability to outperform existing methods. This study also opens up new possibilities for practical applications.
\\ \newline \Keywords{few-shot learning, visually-rich document, variational inference} }

\begin{document}

\maketitleabstract

\section{Introduction}
Relational learning, also known as key-value relation extraction, is a fundamental task in comprehending Visually-Rich Documents (VRDs) \cite{DBLP:conf/das/DengelK02,liu-2019-graph,li-2022-VRD}. It involves automatically identifying and extracting key and value entities, as well as classifying the relations between these entities in a text based on a predefined schema. This task is typically performed on scanned or digitally generated documents, such as invoices, receipts, and business forms.

Recent document AI models pre-trained on large-scale scanned document datasets \cite{xu-2020-layoutlm, hong-2021-BROS, Garncarek-2021-lambert, layoutlmv3-2022-huang} have demonstrated promising performance by effectively leveraging multi-modal information. However, when it comes to real-world applications, the diverse layout formats and form styles found in VRDs present a significant challenge for adapting these AI models \cite{li-2022-VRD}. These approaches lack the inherent ability to automatically detect and identify new types of entities and relations in unfamiliar domains, especially in the absence of annotated data. In contrast, humans possess the remarkable ability to swiftly comprehend key-value patterns in VRD by analyzing just a few lines on the page. 

Despite recent advances \cite{cheng-2020-textfield, wang-shang-2022-towards, wang-2022-formulating}, there is currently no established systematic framework suitable for addressing the more realistic task of few-shot relational learning in VRDs. This research area remains relatively unexplored and presents several unresolved challenges. One of the main challenges is effectively leveraging layout features in few-shot scenarios. In VRDs, key and value entities often have fixed positions and arrangements, such as being located at specific positions and arranged either up-down or left-right. Therefore, incorporating 2D spatial features can provide strong complementary supervision signals when extracting key-value triplets from VRDs. Another challenge lies in the optimal design of the multimodal fusion mechanism for few-shot learning. While it has been demonstrated that multimodal information can greatly enhance supervised learning, determining how to align and aggregate features across different modalities using only a small number of instances remains unclear. These challenges underscore the necessity for further exploration and development in this research direction.



\begin{figure*}[t]
\centering
\includegraphics[width=1.0\linewidth]{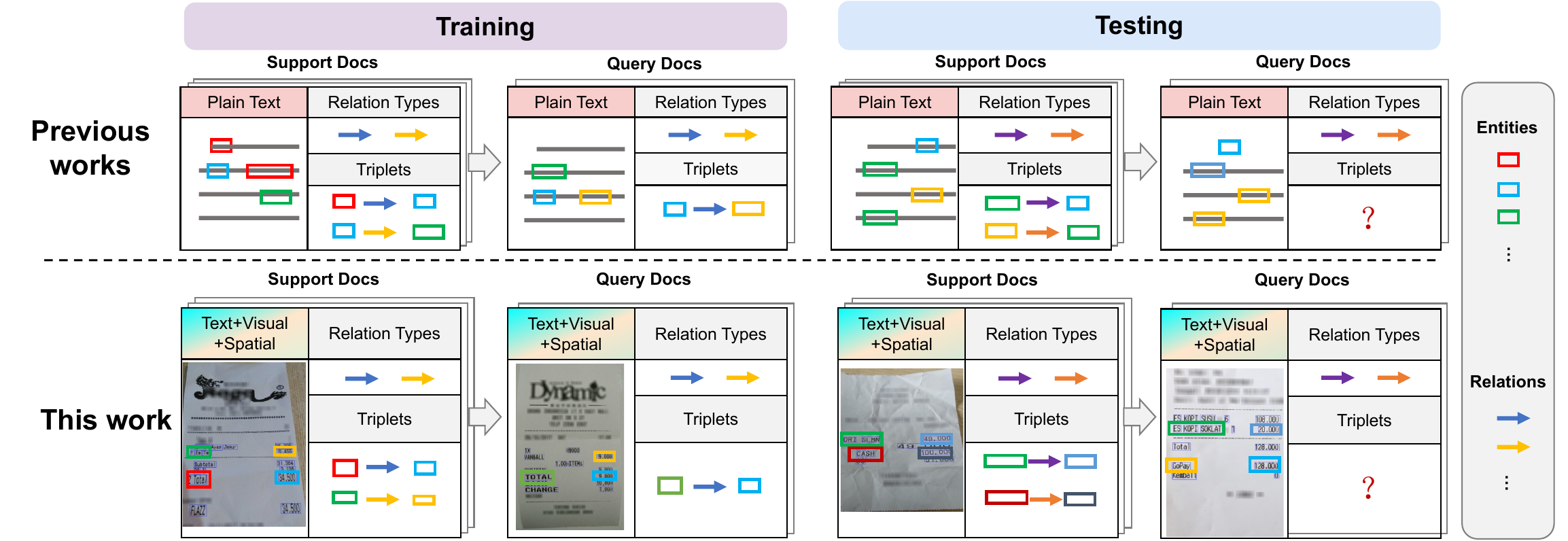}
\caption{Illustration of the distinction between our work and previous works \cite{Popovic2022FewShotDR} during an episode in the few-shot relational learning scenario. In the testing task, we aim to extract triplets that consist of entities and relation types for a given query document. Notably, this task involves a different set of relation types compared to the training task and is performed on a novel collection of documents. Conventional approaches typically rely on off-the-shelf OCR engines to extract text from original document images and solely rely on text features for extracting relational triplets. In contrast, our work takes a human-like perspective and leverages multimodal information to effectively extract the relational triplets. While we use simple receipts with well-aligned layouts to illustrate the idea, it is crucial to acknowledge that real-world scenarios are considerably more complex and challenging.\label{fig:motivation}
}
\end{figure*}







Since there is currently no suitable dataset for the few-shot relational learning task, we first create two benchmark datasets, Few-CORD and Few-SEAB, based on the supervised VRD understanding benchmark datasets CORD \cite{Park-2019-cord} and SEAB \cite{zhang-etal-2022-dualvie}. They allow us to evaluate the model’s ability to transfer knowledge to new classes. Then, we propose a novel variational approach for few-shot relational learning in VRDs to address the challenges mentioned above. This approach incorporates spatial priors and effectively captures multimodal representations, resulting in improved performance on downstream tasks. We utilize a spatial prior encoder to leverage explicit information from Region of Interest (ROI) windows. These windows help us better understand the layout styles in documents and exploit the inherent spatial relationships between entities. To overcome the issue of biased prototypes generated by existing models due to the limited number of $K$-shot instances per category in the training data, we introduce prototypical rectification as a way to access optimal prototypes. In summary, our contributions are as follows:
\begin{itemize} 
\item To the best of our knowledge, this work is the first to tackle the challenge of few-shot relational learning in VRDs. In order to facilitate research in this area, we introduce two new benchmark datasets specifically designed for the few-shot learning setting.
\item We propose a novel variational approach that not only allows for the incorporation of spatial priors but also enables the extraction of robust relation-agnostic features, thereby alleviating the sensitivity of prototypes more accurately.
\item Our method achieves new state-of-the-art performance for few-shot relational learning in VRDs, which has been extensively evaluated on the constructed datasets to demonstrate its effectiveness.
\end{itemize}

\section{Task Description}
As shown in Figure~\ref{fig:motivation}, this task follows the few-shot settings in previous works \cite{han-2018-fewrel,soares-2019-matching} and aims to extract key and value entities involving new relation classes, by training on known relation classes with a small number of examples. The testing set examples often include relation types that were not present in the training set. For practicality and simplicity, we consider only the key and value entities from a single relation within a given document as a training/test example. Then, we adopt a single collapse sequential labeling model \cite{wang-2013-collapsed,islam-2019-collapse} to jointly extract sets of key and value entities ($\mathcal{E}_{k}$ or $\mathcal{E}_{v}$) given their relation types $\mathcal{R}$ from the document $\mathcal{D}$. This is achieved by extending the label space of the previous named entity recognition (NER) scheme for entity types, e.g., ``\texttt{Cash}'' (in CORD) and ``\texttt{Consignee}'' (in SEAB), with additional key or value identifiers (``\texttt{-Key}'', ``\texttt{-Value}'') into the combination labels ``\texttt{Cash-Key/Cash-Value}'' and ``\texttt{Consignee-Key/Consignee-Value}''. In fact, we perform token-level classification to extract entity mentions from the document using a standard ``\texttt{BIO}'' tagging scheme - ``\texttt{Begin, Inside, Other}''. For example, ``\texttt{Consignee-Key-B}'' indicates the first token in the entity mention of ``\texttt{Consignee-Key}'', while ``\texttt{O}'' represents ``\texttt{Other}'' that none of the above applies.

\section{Few-Shot Datasets}

\begin{figure}[t]
\centering
\includegraphics[width=\linewidth]{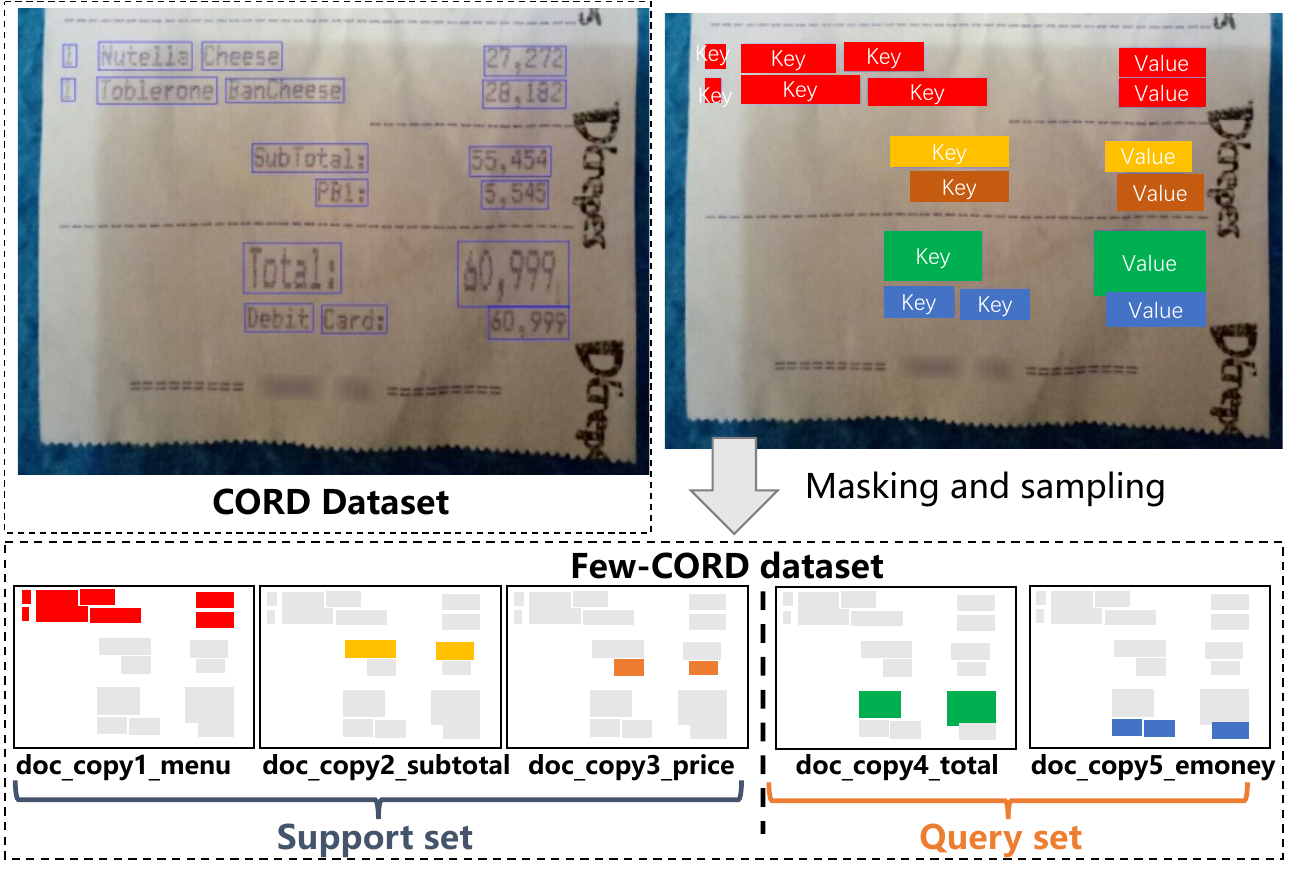}
\caption{Copying, masking and sampling.}
\label{fig:sampling}
\end{figure}

\subsection{Relation-Wise Sampling Strategy} 

During sampling, as shown in Figure~\ref{fig:sampling}, we should carefully select and organize instances from the documents to ensure that the $N$-way setting is maintained while still capturing the necessary contextual information. By implementing an appropriate sampling strategy, our approach aims to strike a balance between incorporating document-level context and adhering to the constraints of the $N$-way setting, enabling effective few-shot learning for sequential labeling tasks. We use the extended dataset consisting of multiple copies of the original documents. We sample from this extended dataset, adding one more converted document to the support or query set after each sampling step until the support or query set reaches the desired number of entity classes ($N$-way) and instances per class ($K$-shot). The overall procedure of the hierarchical sampling method is summarized in Algorithm~\ref{alg:algorithm1}.
 
\begin{algorithm}[t]
\footnotesize
\renewcommand{\algorithmicrequire}{\textbf{Input:}}
\renewcommand{\algorithmicensure}{\textbf{Output:}}
\renewcommand{\algorithmiccomment}[1]{\hfill\textcolor{black!60}{\texttt{$\triangleright$#1}}}
\caption{Relation-wise $N$-way $K$-shot Sampling}
\label{alg:algorithm1}
\begin{algorithmic}[1]
\REQUIRE  {Documents $\hat{\mathcal{D}}$, $N$, $K$, $K^{\prime}$;}
\ENSURE {Support Set $\mathcal{S}$, Query Set $\mathcal{Q}$;} 
\STATE $\mathcal{S} \gets [\; ]$, $\mathcal{Q} \gets [\; ]$;\; \algorithmiccomment{initialization}  
 
\FOR {$j \gets 1$ \TO $N/2$\;\; }
    \STATE $\mathcal{S}[j]\gets \{ \}$;\;\algorithmiccomment{$N/2$ is the number of relation types} 
    \STATE $\mathcal{Q}[j]\gets \{ \}$;\;\algorithmiccomment{$N$ is the number of entity types}
\ENDFOR
\REPEAT 
    \STATE randomly sample $(\mathcal{D}^{(i)}, \mathcal{R}^{(i)})$ from the extended masked document dataset;\; \algorithmiccomment{ $\mathcal{D}^{(i)} \in \hat{\mathcal{D}}$ }
    \IF{ $|\mathcal{S}| < N/2  \And  |\mathcal{S}[j]| < K$ }
         \STATE $\mathcal{S}[j] \gets \mathcal{S}[j] \cup  (\mathcal{D}^{(i)}, \mathcal{R}^{(i)})$; \algorithmiccomment{add to the support set}
    \ENDIF
    \IF{ $|\mathcal{Q}| < N/2  \And  |\mathcal{Q}[j]| < K^{\prime}$ and $(\mathcal{D}^{(i)}, \mathcal{R}^{(i)}) \notin \mathcal{S}[j]$ }
        \STATE $\mathcal{Q}[j] \gets \mathcal{Q}[j] \cup (\mathcal{D}^{(i)}, \mathcal{R}^{(i)})$; \algorithmiccomment{add to the query set}
    \ENDIF
\UNTIL {$|\mathcal{S}|=|\mathcal{Q}|=N/2$ \AND  $\{ \forall j \mid |\mathcal{S}[j]|= K$ \AND $|\mathcal{Q}[j]|= K^{\prime}\}$}
\RETURN $\mathcal{S}$, $\mathcal{Q}$;
\end{algorithmic}
\end{algorithm}
In our approach, we employ $N$-way\footnote{Note that $N/2$ denotes the number of relation types thus $N$ denotes the number of entity types (key+value).} $K$-shot learning to train our few-shot relational learning system. This involves constructing training or testing episodes iteratively. While considering the document-level context is important for few-shot learning tasks, sampling at the document level can pose challenges within the $N$-way setting. This is because documents often contain multiple relations/entities, which may exceed the limits of the $N$-way setting. To maintain the $N$-way setting, as shown in Figure~\ref{fig:sampling}, we hereby make relation-wise copies for each document which means each copy contains only entities involved in a particular relation type, and other bounding boxes are simply relabeled as \texttt{Other} type. The mathematical expression is: $
\mathcal{D}_\mathcal{R} = \mathcal{M}(\mathcal{D}, \bar{\mathcal{R}})
$, where $\mathcal{D}_\mathcal{R}$ represents the copy in which irrelevant entities to $\mathcal{R}$ are masked. The masking operation $\mathcal{M}$ selectively retains only the entities related to $\mathcal{R}$ and suppresses or eliminates the rest, denoted as $\bar{\mathcal{R}}$. 

For each episode, we randomly choose $N$ classes ($N$-way) and sample $K$ examples ($K$-shot) from the extended dataset containing multiple copies for each relation class to build support set $\mathcal{S}_{train}=\left\{\mathcal{D}^{(i)}, \mathcal{R}^{(i)}\right\}_{i=1}^{N \times K}$ and $K^{\prime}$ examples for a query set $\mathcal{Q}_{train}=\left\{\mathcal{D}^{(j)}, \mathcal{R}^{(j)}\right\}_{j=1}^{N\times K^{\prime}}$, ensuring that the support set $\mathcal{S}$ and query set $\mathcal{Q}$ do not overlap ($\mathcal{S} \cap \mathcal{Q}=\emptyset$). In addition, when adding one or more examples to the support or query set after each sampling step until the support or query set reaches the desired number of entity classes ($N$-way) and instances per class ($K$-shot). 

In the training phase, we train the few-shot learning system using the support and query sets ($\mathcal{S}_{train}$, $\mathcal{Q}_{train}$), where the supervisions for both sets are visible. In the testing phase, we predict the new classes in the query set $\mathcal{Q}_{test}$ and evaluate the performance given the ground truths.



\begin{figure*}[htbp]   
  \begin{minipage}[t]{0.24\linewidth}
		\centering
		\includegraphics[width=1.7in]{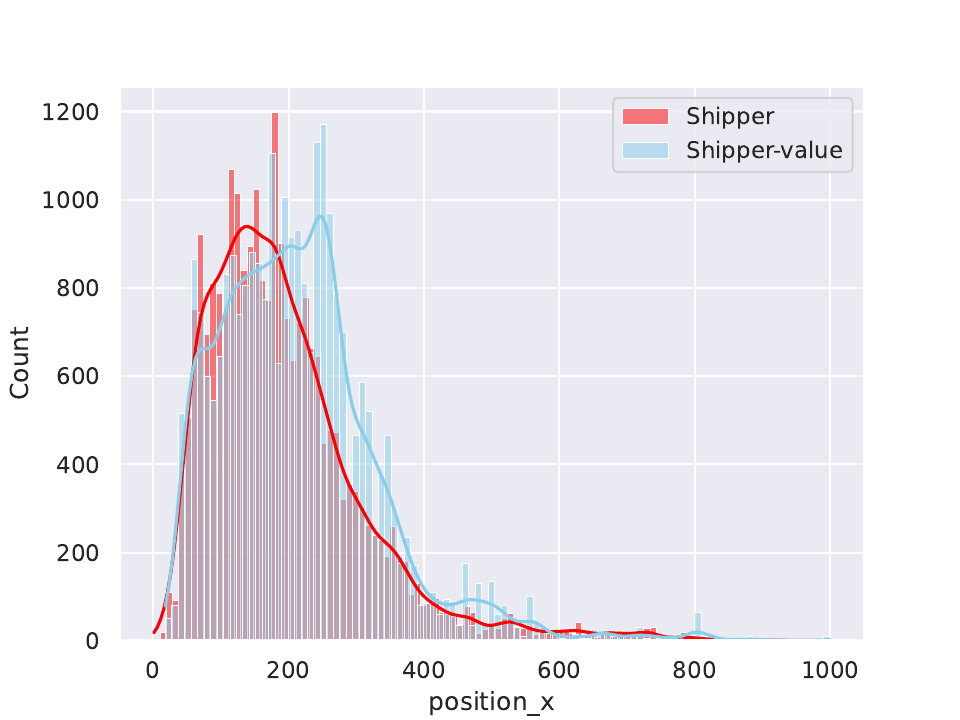}
	\end{minipage}
	\begin{minipage}[t]{0.24\linewidth}
		\centering
		\includegraphics[width=1.7in]{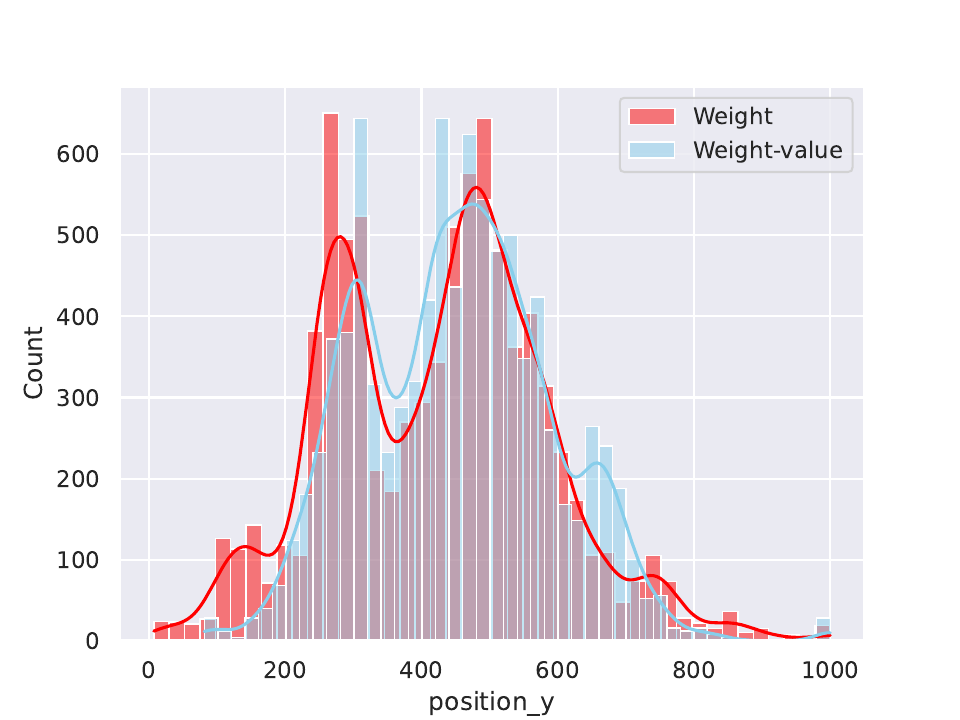}
	\end{minipage}
    \begin{minipage}[t]{0.24\linewidth}
		\centering
		\includegraphics[width=1.7in]{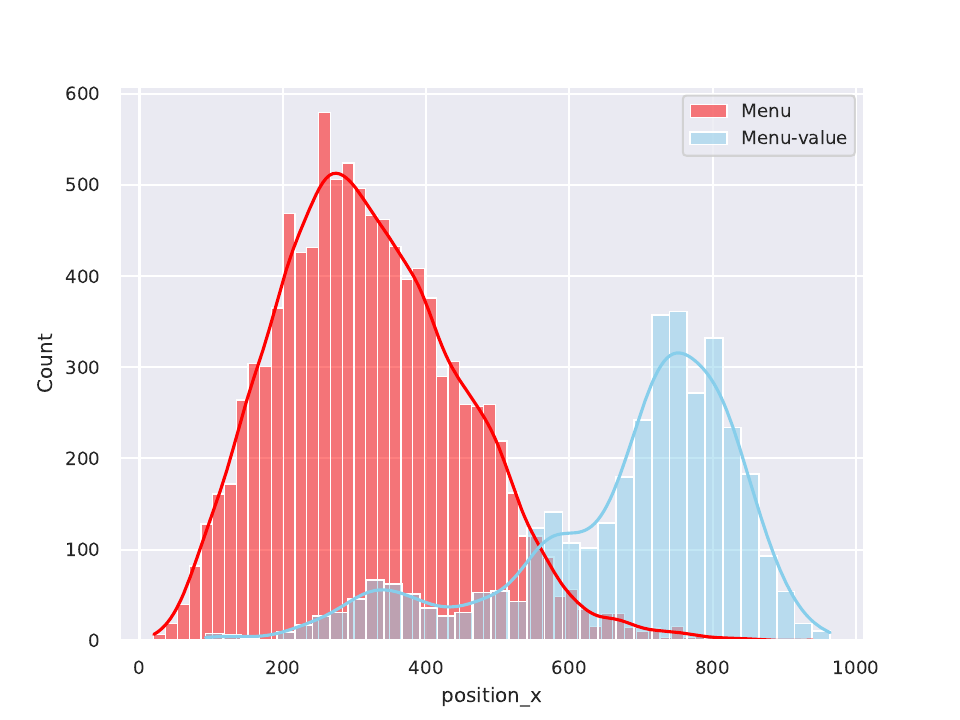}
	\end{minipage}
    \begin{minipage}[t]{0.24\linewidth}
		\centering
		\includegraphics[width=1.7in]{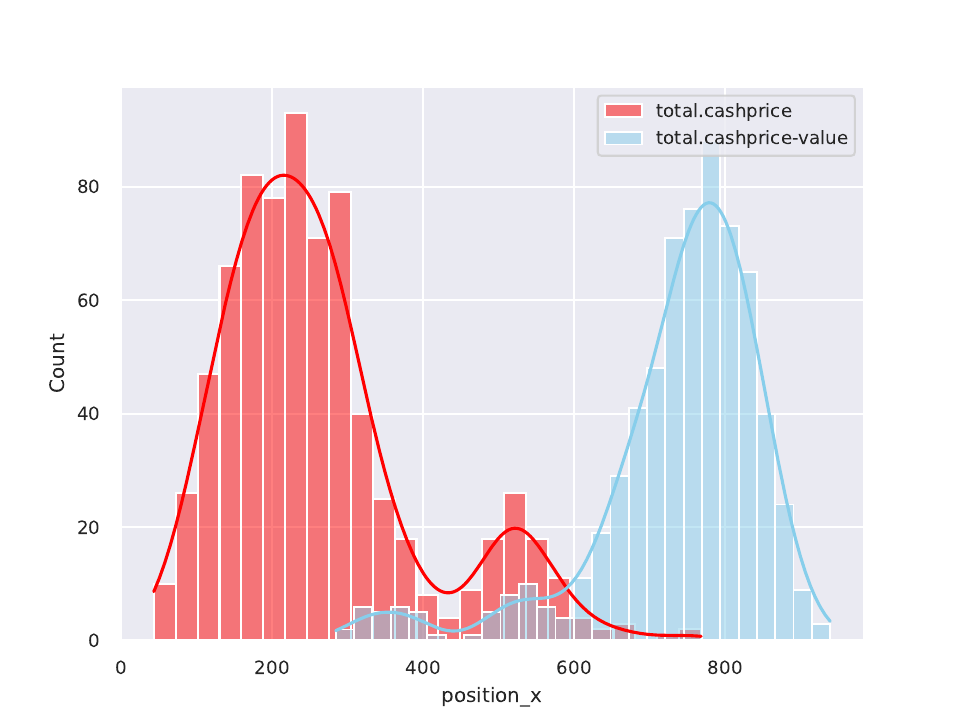}
	\end{minipage}
	\caption{The double-humped distribution of specific key-value types on a document page, including: ``\texttt{Shipper}'' and ``\texttt{Weight}'' in the SEAB dataset and ``\texttt{Menu}'' and ``\texttt{Total}'' in the CORD dataset.} 
	\label{fig:hump} 
\end{figure*}
\subsection{Dataset Details}
This section provides an overview of the Few-CORD and Few-SEAB datasets. We construct these two benckmark datasets on the top of the CORD \cite{Park-2019-cord} and SEAB \cite{zhang-etal-2022-dualvie} datasets and serve for few-shot relational learning. These datasets consist of diverse and realistic examples, enabling us to effectively measure the effectiveness of relational learning from VRDs. Table~\ref{table_data} provides the details of the newly created few-shot datasets.

\begin{table}[t]
\begin{center}

\scalebox{0.65}{
\begin{tabular}{lrrrcrrr}
\toprule
\multirow{2}{*}{\textbf{Dateset}} & \multicolumn{3}{c}{\textbf{Train}}& & \multicolumn{3}{c}{\textbf{Test}} \\ \cmidrule{2-4} \cmidrule{6-8}
&  \#Doc     & \multicolumn{1}{c} \#BD   &   \#Types  &  & \#Doc      & \multicolumn{1}{c}{ \#BD }      &  \#Types      \\ \midrule
CORD                             & 800     & 18,915  & 32/16   &  & 200       & 4,466      & 32/16     \\ 
SEAB                          & 3,562    & 249,255 & 44/22   & & 953      & 73,873      & 44/22    \\ \midrule
Few-CORD                             & 1,211     & 32,160  & 18/9   & & 702      & 15928      & 14/7     \\ 
Few-SEAB                          & 20,831     & 575,391  & 24/12   & & 4,048      & 146,312      & 20/10      \\ 
 \bottomrule
\end{tabular}
}
\caption{
Statistics of the supervised and few-shot datasets, including the numbers of documents (Doc), bounding boxes (BD), and types of entities/relations (Types A/B means A: base classes in training; B: novel classes in testing).
}
\label{table_data}
 \end{center}
\end{table} 

\begin{figure*}[t]
\centering
\includegraphics[width=1.0\textwidth]{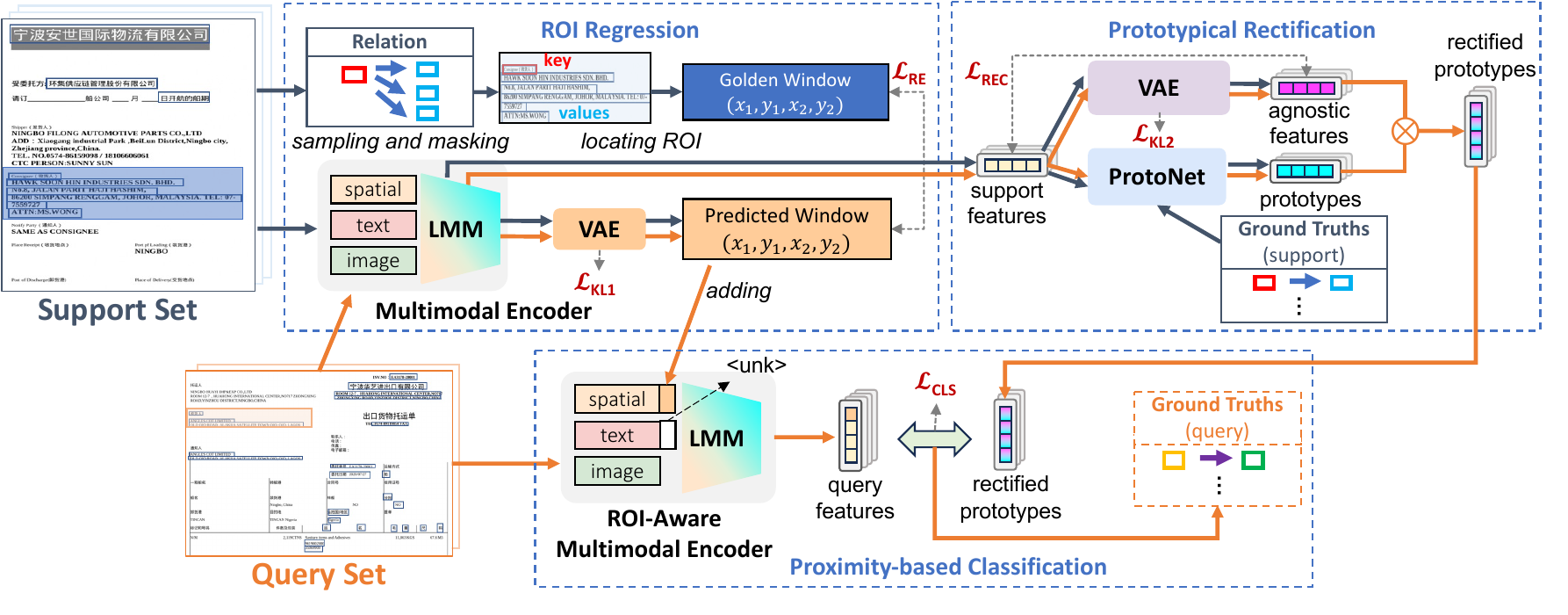}
\caption{Our model architecture comprises three key components: ROI regression, prototypical rectification, and proximity-based classification. It can
generate more robust representations that encompass multiple modalities by directing its attention to relevant regions (by predicting the explicit ROI windows) and learn high-dimensional relation-agnostic features using prototypical rectification, helping adaptation to new relation classes.
}
\label{fig:my_model}
\end{figure*}


\section{Human-Like Approaches} 
\subsection{Motivation} 
People possess the remarkable ability to quickly grasp relation patterns with minimal exposure to instances. They can effectively utilize layout information across various document images, given the explicit spatial relationships between key and value entities. These two-dimensional spatial priors serve as valuable cues for learning of the relations in VRDs. However, existing approaches struggle to effectively leverage these features. Moreover, people have the capacity to infer insights from unseen examples after seeing only a few instances. They can learn high-dimensional class-agnostic features that transcend linguistic boundaries \cite{wang-2022-lilt} and contextual limitations \cite{Han2023FewShotOD} by establishing connections across different classes, irrespective of their novelty or familiarity. This unique capability allow them to comprehend and generalize knowledge within diverse contexts to identify new class relations and entities. 

Therefore, in this study, our objective is to introduce a novel approach that emulates human-like behavior in  learning relations. Our approach focuses on incorporating relational 2D-spatial priors and aims to learn high-dimensional class features, irrespective of their novelty or familiarity. By combining these two aspects, we aim to bridge the gap between human cognition and existing models, thereby enabling more effective few-shot relational learning in VRDs.

\subsection{Architecture} 
Figure~\ref{fig:my_model} illustrates the overall architecture of our method, which comprises three stages: 1) \textbf{ROI Regression}, which learn regions of interest (ROIs) using carefully curated golden window instances; 2) \textbf{Prototypical Rectification}, which tackles the issues associated with biased prototype representations; and 3) \textbf{Proximity-based classification}, which predicts the final token labels.
Figure~\ref{fig:my_model} visually illustrates the comprehensive architecture of our method, which involves three key stages.
 

\subsection{ROI Regression}


Our preliminary studies have revealed a strong correlation between specific key and value entities in the two-dimensional layout space of VRDs. In general, keys and values tend to exhibit a distinctive double-hump distribution within this spatial arrangement (see Figure~\ref{fig:hump}), which can be seen as a de facto two-dimensional spatial prior.

To leverage this distributional regularity, we introduce the concept of a ``golden window'' that encompasses both the keys and values belonging to a particular type. During training, this golden window serves as an explicit supervision signal, guiding the model to focus on relevant local regions of interest (ROIs). We first employ LayoutLM \cite{xu-2020-layoutlm} and LayoutLMv2 \cite{xu-2021-layoutlmv2} as our multimodal encoder to extract token-level features $\mathbf{H}$ from multimodal channels, including text, spatial (1D position and 2D layout), and visual. For simplicity, we call them LLM. Then, we utilize a Variational Autoencoder (VAE) \cite{kingma-2013-Bayes} to model the distributions of local ROIs. Specifically, the output of the decoder in VAE corresponds to four dimensions $(x_1, y_1,x_2,y_2)$ used for reconstructing the golden window. This reconstruction enables improved accuracy in predicting key-value associations by explicitly considering their geometric layout within visually-rich documents.

\subsection{Prototypical Rectification}
Given the support features extracted by the multimodal encoder, we utilize the prototypical network (ProtoNet) paradigm \cite{snell-2017-protonet} to compute an M-dimensional representation $\mathbf{p}_{c} \in \mathbb{R}^M$ for each entity class. We take the average of the token embeddings to obtain prototypes. 

However, ProtoNet encodes the support instances as a single feature vector, but this does not provide an accurate estimate of the class center \cite{yang-2020-enhanceproto,gao-2019-hybrid,liu-2022-learn}, especially when the data is sparse, and the examples are highly variable, making it difficult to represent the distribution of the class.

Building on recent advancements in variational feature learning \cite{Han2023FewShotOD,xia-2020-variational,vk-2018-advcross}, we introduce a variational rectification mechanism to incorporate category distribution information into the prototypes. This mechanism enables us to transform support features into class-wise representations. We expect the rectified prototypes $\tilde{\mathbf{z}}_c$ to capture more generic features of the class that are robust to the variance of support instances. Here, $\mathbf{p}_c$ represents the prototype feature of class $k$. We approximate the class distribution $N(\mu_{\rm{p}}, \Sigma_{\rm{p}})$ and sample a variational feature $\tilde{\mathbf{z}}_{\rm{p}}$ from this distribution. Then, we combine the prototypes and the variational feature using the following equation:
\begin{equation}
\tilde{\mathbf{p}}_{c}=\mathcal{A}(\mathbf{p}_{c}, \tilde{\mathbf{z}}_{\rm{p}})=\mathbf{p}_{c} \otimes  sigmoid(\tilde{\mathbf{z}}_{\rm{p}}), c \in C
\end{equation}

\subsection{Proximity-based Classification}
To incorporate the ROI information, we enhance the multimodal input sequence by appending the coordinates of the predicted window ${b}_{\rm{pred}}$ to the end of the sequence, specifically for the spatial modality. In order to maintain alignment between the multimodal inputs, we introduce a special token $\mbox{\texttt{<UNK>}}$ to the corresponding text modality. Consequently, the resulting incorporated sequence is extended from the initial $n$ tokens to $n+1$ tokens, denoted as $[t_1, t_2, \dots, t_n, \mbox{\texttt{<UNK>}}]$. We take the last hidden layer outputs of the ROI-aware multimodal encoder $\tilde{\mathbf{H}}$ as the final query features. This embedding captures sufficient multimodal information while incorporating the ROI information for better relational representations.
\begin{equation}
\tilde{\mathbf{H}}  =  \mbox{LMM}
\Big(
\begin{bmatrix}
t_1, &t_2 , & \dots & t_n, &  \mbox{\texttt{<UNK>}} \\ 
b_1, &b_2, &\dots &b_n,  & b_{\rm{pred}} \\  
\mathcal{I}_{b_1}, &\mathcal{I}_{b_2}, &\dots &\mathcal{I}_{b_n},    & \mathcal{I}_{b_{\rm{pred}}}
\end{bmatrix}\Big)
\end{equation}


For each query instance $\mathcal{Q}_{\rm{test}}$, we perform a proximity-based classification by computing the 2-norm distance (i.e., the Euclidean distance) between the query embeddings $\tilde{\mathbf{h}}_i$ and rectified prototypes $\tilde{\mathbf{p}}_{c}$.
\begin{equation}
\mathbf{d}_{c}=d_{\rm{l2-norm}}(\mathbf{p}_{c},\tilde{\mathbf{h}}_i).
\end{equation}
Then, we normalize the prediction probability of $x$ over all classes using the softmax function. The model predicts the label as the nearest prototype to the input token. 
The cross-entropy loss is used as the loss function for multi-classification task. 

\begin{table*}[t]
\centering
\scalebox{0.7}{
\begin{tabular}{llcccccccccccccc}
\toprule 
\multicolumn{1}{l}{\multirow{2}{*}{\textbf{Method}}} & \multicolumn{1}{l}{\multirow{2}{*}{\textbf{LLM}}} & \multicolumn{1}{c}{\multirow{2}{*}{\textbf{Proto}}}& \multicolumn{1}{c}{\multirow{2}{*}{\textbf{VAE}}}&\multicolumn{5}{c}{\textbf{Few-CORD}} && \multicolumn{5}{c}{\textbf{Few-SEAB}}\\ \cmidrule{5-9} \cmidrule{11-15} 
\multicolumn{1}{c}{}         &   & &        & \textsc{1-shot} & \textsc{2-shot} & \textsc{3-shot} & \textsc{4-shot} & \textsc{5-shot} & &\textsc{1-shot} & \textsc{2-shot} & \textsc{3-shot} & \textsc{4-shot} & \textsc{5-shot}\\ \midrule
ProtoNet & BERT                      & \Checkmark & \XSolidBrush            & 32.32   &  \underline{35.14}   & \underline{38.87}   & \underline{40.08}   & \underline{42.95}  & & 26.88  & 28.59  & \underline{30.08}  & \underline{31.74}  &\underline{34.11}\\ 
NNShot & BERT            & \Checkmark& \XSolidBrush                       & 29.19   & 32.34   & 35.16   & 36.28   & 38.04  & & 25.16  & 26.10  & 27.92  & 28.80  & 29.95\\ 
StructShot & BERT       &\Checkmark & \XSolidBrush                              & \textbf{33.54}   & 34.95   & 37.41   & 38.31   & 40.38  & &\underline{27.30}  & \underline{28.63}  & 29.20  & 30.19  & 31.75\\ 
VFA & BERT       & \XSolidBrush& \Checkmark                              & 30.08   & 31.34   & 32.86   & 34.89   & 37.14  & &24.83  & 25.61  & 27.11  & 28.65  & 29.06\\ 
\rowcolor{gray!20} \textbf{ProtoRec} & BERT         & \Checkmark & \Checkmark                         &  \underline{33.30}   & \textbf{35.40}   & \textbf{39.01}   & \textbf{40.53}   & \textbf{43.14}  & & \textbf{27.45}  & \textbf{29.04}  & \textbf{30.35}  & \textbf{32.15}  & \textbf{34.28}\\ \midrule
ProtoNet & LayoutLM        & \Checkmark& \XSolidBrush                          & 70.25   & 74.10   & 77.02   & 79.31   & 80.40  & & 60.95  & 64.02  & 66.31  & 69.53  & 73.12\\ 
NNShot & LayoutLM    & \Checkmark& \XSolidBrush                             & 68.20   & 72.70   & 73.76   & 75.24   & 76.67  & & 58.80  & 61.75  & 62.27  & 64.40  & 66.89\\ 
StructShot & LayoutLM      & \Checkmark& \XSolidBrush                           & \underline{71.38}   & 73.88   & 74.52   & 77.24   & 77.83  & & 61.15  & 63.14  & 63.50  & 65.28  & 68.10\\ 
VFA & LayoutLM      & \XSolidBrush& \Checkmark                           & 68.39   & 69.79   & 71.83   & 73.18   & 74.98  & & 58.14  & 60.23  & 63.14  & 64.36  & 66.12\\ 
VFA+ROI & LayoutLM      & \XSolidBrush& \Checkmark                           & 68.71   & 70.07   & 72.31   & 73.91   & 75.65  & & 58.76  & 60.61  & 63.91  & 65.08  & 66.79\\ 
\rowcolor{gray!20} \textbf{ROI-Aware}  & LayoutLM      & \Checkmark& \Checkmark                           & 71.33   & \underline{74.96}   & \underline{77.84}   & \underline{80.78}   & \underline{81.22}  & & \underline{61.79}  & \underline{64.92}  & \underline{66.45}  & \underline{69.84}  & \underline{73.19}\\ 
\rowcolor{gray!20} \textbf{ProtoRec+ROI} & LayoutLM       & \Checkmark& \Checkmark                         & \textbf{73.21}      & \textbf{76.19}      & \textbf{78.42}      &\textbf{81.35}      &\textbf{81.54} & & \textbf{62.77}  & \textbf{65.54}  & \textbf{66.59}  & \textbf{69.95}  & \textbf{73.28}     \\ \midrule
ProtoNet & LayoutLMv2    & \Checkmark& \XSolidBrush                  & 70.30    & 74.22    & 77.16    & 79.35    & 80.52 & & 61.18      & 64.10      & 66.43      & 69.80      & 73.37    \\ 
NNShot  &LayoutLMv2    & \Checkmark& \XSolidBrush                & 68.17    & 72.80    & 73.88    & 75.39    & 76.84 & & 59.09      & 61.81      & 62.35      & 64.49      & 67.14    \\ 
StructShot & LayoutLMv2    & \Checkmark& \XSolidBrush                & 71.45    & 73.95    & 74.62    & 77.50    & 78.11 & & 61.30      & 63.34      & 63.57      & 65.37      & 68.18    \\ 
VFA & LayoutLMv2    & \XSolidBrush& \Checkmark                & 68.87    & 70.16    & 72.05    & 73.63    & 75.56 & & 58.77      & 60.59      & 63.70      & 65.03      & 67.18    \\ 
VFA+ROI & LayoutLMv2    & \XSolidBrush& \Checkmark                & 69.21    & 71.39    & 72.63    & 74.15    & 76.01 & & 59.20      & 61.57      & 63.97      & 65.28      & 68.26    \\ 
\rowcolor{gray!20}\textbf{ROI-Aware} & LayoutLMv2      & \Checkmark& \Checkmark                  & \underline{71.59}    & \underline{75.83}    & \underline{77.87}    & \underline{80.92}    & \underline{81.40} & & \underline{61.94}      & \underline{65.02}     & \underline{66.48}      & \underline{69.86}      & \underline{73.52}    \\ 
\rowcolor{gray!20}\textbf{ProtoRec+ROI} & LayoutLMv2    & \Checkmark& \Checkmark                 & \textbf{73.37}    & \textbf{76.90}    & \textbf{78.54}    & \textbf{81.46}    & \textbf{81.88} & & \textbf{63.05}  & \textbf{65.59}      & \textbf{66.83}      & \textbf{70.18}      & \textbf{73.70}    \\ \midrule
\rowcolor{gray!20}\textbf{ProtoRec+ROI+CF}& LayoutLMv2   & \Checkmark& \Checkmark     & \underline{73.32}    & \underline{76.96}    & \underline{78.63}    & \underline{81.40}    & \underline{81.85} & & \underline{62.98}      & \underline{65.47}      & \underline{66.80}      & \underline{70.21}      & \underline{73.65}    \\    
\rowcolor{gray!20}\textbf{ProtoRec+ROI+SW} & LayoutLMv2  & \Checkmark& \Checkmark      & \textbf{73.56}    & \textbf{77.39}    & \textbf{78.80}    & \textbf{81.88}    & \textbf{82.18} & & \textbf{63.43}      & \textbf{65.87}      & \textbf{67.30}      & \textbf{70.56}     & \textbf{73.96}   \\
\bottomrule
\end{tabular}}
\caption{\label{cord-and-seab}
The averaged f1-scores on the Few-CORD and Few-SEAB datasets. \textbf{Bold} and \underline{underline} indicate the best and second-best scores in each group. \textbf{ROI-Aware} and \textbf{ProtoRec} are two main components of the variational model proposed in this work. \textbf{+CF} denotes filling the windows with a light color. \textbf{+SW} refers to adversarial learning 
 by shrinking windows. }
\end{table*}

\subsection{Training Objective}
We train our few-shot relation learning model in an end-to-end manner with the following multi-task loss,
\begin{equation}
\mathcal{L}_{\rm{final}} =\mathcal{{L}}_{\rm{REC}} +   \alpha \mathcal{{L}}_{\rm{KL1}}+\mathcal{{L}}_{\rm{RE}} + \beta \mathcal{{L}}_{\rm{KL2}} + \mathcal{L}_{\rm{CLS}},
\end{equation}
where the weight coefficients ${\alpha}=\beta=2.5\times10^{-4}$.

In the training phase, parameters are updated in each episode. In the testing phase, we directly deploy the model to predict novel types of entities without computing the loss or updating the model parameters.

\section{Implementation Details}
\subsection{Backbones}
We implement our model based on the approach proposed by \newcite{ding-2021-fewnerd}. For the backbone architecture, we utilize pre-trained document intelligence models such as LayoutLM \cite{xu-2020-layoutlm} and LayoutLMv2 \cite{xu-2021-layoutlmv2}. These models serve as the foundation for our method, providing robust and effective feature extraction capabilities. In addition, we also evaluate the performance of the baseline model using Bert \cite{bert-2018-Devlin} solely for the text modality. 
\subsection{Hyper-parameters}
We optimize the model with AdamW optimizer on a dual NVIDIA 3090 GPUs machine, and the learning rate is $1\times10^{-5}$. We fine-tune the model with 10,000 iterations in training and evaluate the performance using the averaged scores over 500 testing iterations.
\subsection{Evaluation Scheme}
Following \cite{han-2018-fewrel,gao-2019-fewrel,ding-2021-fewnerd}, we adopt episode evaluation. We compute the micro-F1 score over a number of test episodes. Each episode contains a $K$-shot support set with manually annotated labels and a $K^{\prime}$-shot query set without any annotated labels.
\subsection{Baselines}
\textbf{ProtoNet}: is a baseline prototype system that assigns each token representation to the nearest label representation by learning the examples in the training set using a prototypical network \cite{snell-2017-protonet,garcia-2018-fewshot}. 

\textbf{NNShot} and \textbf{StructShot} \cite{yang-2020-nnshot}: are state-of-the-art methods based on token-level nearest neighbor classification. In contrast to the ProtoNet, NNShot determines the tag of a query based on token-level distance. On the other hand, StructShot incorporates an additional Viterbi decoding during the inference phase to improve the overall performance.

\textbf{VFA}: Variational Feature Aggregation (VFA) \cite{Han2023FewShotOD} is originally proposed for few-shot target detection in the CV domain. We re-implement and adapt it to our specific task. 



\section{Experiment Results}
\subsection{Overall Evaluation}

Table~\ref{cord-and-seab} shows the f1-scores on the Few-CORD and Few-SEAB datasets. In general, we have observed that the BERT backbone performs much worse than LayoutLM and LayoutLMv2. This is primarily due to BERT's limitation in capturing non-textual information. In contrast, LayoutLM and LayoutLMv2, which includes visual information, proves to be more effective. However, since LayoutLM itself does not inherently model visual features, we utilize ResNet as its visual encoder. Nonetheless, the misalignment between visual modalities and other modalities affects the effectiveness of LayoutLM as compared to LayoutLMv2.

From a horizontal perspective, it is generally observed that as the number of shots, denoted as $K$, increases, the performance of all methods tends to improve. This improvement can be attributed to the fact that as $K$ increases, the prototypical representations become closer to the actual data points, leading to better performance. Overall, our prototype rectification (\textbf{ProtoRec}) and \textbf{ROI-aware} methods consistently yield stable improvements regardless of the backbone model used. It is worth mentioning that the combination of ProtoRec and ROI-Aware (\textbf{ProtoRec+ROI-Aware}) achieves the best performance among the methods mentioned. It is important to note that adding a Layout-aware Encoder to BERT's backbone is not possible due to BERT's lack of 2D position information, so we only show the ProtoRec result in the BERT group.
\subsection{Semantic Similarity}
\begin{figure}[t]
\centering
\includegraphics[width=0.5\textwidth]{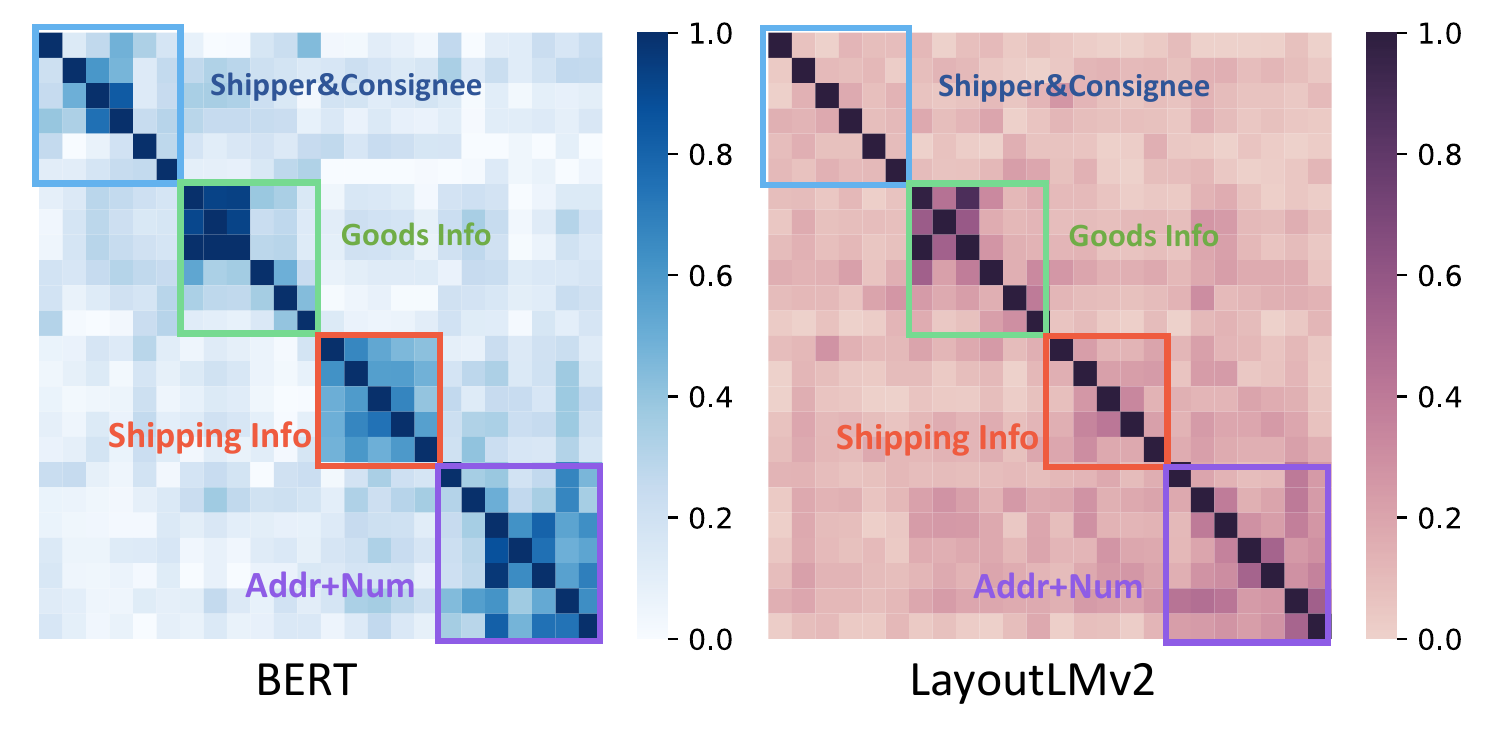}
\caption{Semantic similarity heatmap for the entity representations generated by using BERT and LayoutLMv2.}
\label{similarity1}
\end{figure}

Prototype sharing \cite{R-2021-pshare} can help identify prototypical similarities between classes. In this study, we aim to explore the relevance of relations and semantic similarity between entities, conducting an empirical study on entity-class similarity. For this purpose, we employ BERT and LayoutLMv2, having pre-trained on the training set to generate embeddings for all entities mentioned in the test set. We randomly selected 100 instances of entity embeddings for each fine-grained type and averaged them. Then, we calculate the dot product between the central representations of each entity type to measure their similarity. By comparing the results obtained from BERT with those from LayoutLMv2, we gain insights into the similarity. Figure~\ref{similarity1} depicts that entity types sharing the same coarse-grained type tend to exhibit greater similarity, thereby facilitating knowledge transfer. 

\begin{table*}[t]
\centering
\scalebox{0.72}{
\begin{tabular}{llcccccccccccccc}
\toprule
\multicolumn{1}{l}{\multirow{2}{*}{\textbf{Method}}} & \multicolumn{1}{l}{\multirow{2}{*}{\textbf{LLM}}} & \multicolumn{1}{c}{\multirow{2}{*}{\textbf{Proto}}}& \multicolumn{1}{c}{\multirow{2}{*}{\textbf{VAE}}}&\multicolumn{5}{c}{\textbf{Inter}} && \multicolumn{5}{c}{\textbf{Intra}}\\ \cmidrule{5-9} \cmidrule{11-15} 
\multicolumn{1}{c}{}         &  & &         & \textsc{1-shot} & \textsc{2-shot} & \textsc{3-shot} & \textsc{4-shot} & \textsc{5-shot} & &\textsc{1-shot} & \textsc{2-shot} & \textsc{3-shot} & \textsc{4-shot} & \textsc{5-shot}\\ \midrule
ProtoNet & BERT  & \Checkmark & \XSolidBrush      & 33.03      & \underline{36.73}      & \underline{37.08}      & \underline{38.36}      & \underline{39.54}    & & 24.39      & \underline{26.68}      & \underline{27.86}      & \underline{28.35}      & \underline{29.87}\\ 
NNShot & BERT  & \Checkmark & \XSolidBrush        & 34.35      & 34.68      & 35.75      & 35.93      & 36.47    & & 23.46      & 24.43      & 25.10      & 25.57     & 26.20\\ 
StructShot & BERT  & \Checkmark & \XSolidBrush        & \textbf{36.45}      & 36.63      & 36.83      & 37.10      & 37.29    & & \textbf{25.28}      & 25.87      & 26.19      & 26.86     & 27.22\\ 
VFA & BERT    & \XSolidBrush & \Checkmark                                 & 31.65   & 33.87   & 35.48   & 36.11   & 36.69 &  & 21.90& 23.31 & 24.58  & 25.42  & 25.94  \\ 
\rowcolor{gray!20}\textbf{ProtoRec} & BERT   & \Checkmark & \Checkmark            & \underline{34.80}      & \textbf{37.10}      & \textbf{37.45}      & \textbf{38.71}      & \textbf{39.86}    & & \underline{25.06}      & \textbf{27.18}      & \textbf{28.51}      & \textbf{28.92}     & \textbf{30.09}\\ \midrule
ProtoNet & LayoutLM & \Checkmark & \XSolidBrush    & 66.75      & 69.31      & 70.86      & 74.88      & 77.84    & & 55.67      & 58.23      & 60.92      & 64.20     & 65.89\\ 
NNShot & LayoutLM & \Checkmark & \XSolidBrush   & 63.83      & 65.50      & 66.32      & 68.43      & 70.88    & & 53.74      & 56.37      & 58.83      & 62.13     & 63.48\\ 
StructShot & LayoutLM & \Checkmark & \XSolidBrush   & 67.69      & 69.35      & 69.64      & 71.64      & 72.35    & & 55.53      & 57.90      & 58.96      & 62.41     & 64.81\\ 
VFA & LayoutLM   & \XSolidBrush & \Checkmark                                & 64.45   & 65.89   & 67.34   & 70.63   & 72.90 &  & 51.48 & 54.67 & 55.86  & 59.34  & 61.22  \\ 
VFA+ROI & LayoutLM       & \XSolidBrush & \Checkmark                                & 65.32   & 66.17   & 67.81   & 71.23   & 73.12  & & 51.96 & 55.04  & 56.12  & 59.76  & 61.46\\ 
\rowcolor{gray!20}\textbf{ROI-Aware} & LayoutLM  & \Checkmark & \Checkmark            & \underline{68.80}      & \underline{70.39}      & \underline{71.48}      & \underline{75.60}      & \underline{78.17}    & & \underline{56.80}      & \underline{58.31}      & \underline{61.20}      & \underline{64.38}     & \underline{66.95}\\ 
\rowcolor{gray!20} \textbf{ProtoRec+ROI} & LayoutLM   & \Checkmark & \Checkmark      & \textbf{69.67}      & \textbf{71.18}      & \textbf{71.88}      & \textbf{75.83}      & \textbf{78.64}   
 & & \textbf{57.94}      & \textbf{59.13}      & \textbf{61.79}      & \textbf{64.85}     & \textbf{67.05} \\ \midrule
ProtoNet & LayoutLMv2 & \Checkmark & \XSolidBrush   & 67.21      & 69.55      & 71.23      & 75.14      & 78.22  
 & & 56.16      & 58.52      & 61.65      & \underline{64.96}      & 66.31\\ 
NNShot & LayoutLMv2& \Checkmark & \XSolidBrush & 63.87      & 66.20      & 66.87      & 68.87      & 71.26  
 & & 54.70      & 57.23      & 58.93      & 62.27      & 63.02\\ 
StructShot & LayoutLMv2 & \Checkmark & \XSolidBrush & 68.16      & 69.73      & 70.75      & 71.63      & 74.27  
 & & 56.82      & 58.24      & 59.21      & 62.54      & 63.27\\ 
VFA & LayoutLMv2      & \XSolidBrush & \Checkmark                            & 64.89   & 66.46   & 68.05   & 70.98   & 73.59 &  & 52.42& 55.61 & 56.74  & 60.38  & 61.69  \\ 
VFA+ROI & LayoutLMv2      & \XSolidBrush & \Checkmark                             & 65.43   & 67.04   & 68.78   & 71.61   & 74.04  & & 52.95 & 56.38  & 57.61  & 60.84  & 61.93\\ 
\rowcolor{gray!20}\textbf{ROI-Aware} & LayoutLMv2 & \Checkmark & \Checkmark     & \underline{70.11}      & \underline{71.19}      & \underline{72.64}      & \underline{75.80}      & \underline{78.67}    & & \underline{57.69}      & \underline{59.80}      & \underline{61.91}      & 64.87      & \underline{67.14}\\
\rowcolor{gray!20}
\textbf{ProtoRec+ROI} & LayoutLMv2 & \Checkmark & \Checkmark & \textbf{71.59}      & \textbf{72.76}      & \textbf{73.32}      &\textbf{75.97}      & \textbf{79.59}   & & \textbf{58.79}      & \textbf{60.44}      & \textbf{62.37}      & \textbf{65.13}      & \textbf{67.53}\\ \bottomrule
\end{tabular}
}
\caption{\label{inter-intra}
The averaged f1-score on the Few-SEAB (Inter) and Few-SEAB (Intra) datasets. Few-SEAB (Inter) requires the entity types to be disjoint in the training set and the test set. Few-SEAB (Intra) requires the entity types to be disjoint in the training set and the test set, while Few-SEAB (Inter) shares the coarse-grained types in the training and test sets.}
\end{table*}

\subsection{Transfer Learning Capacity}
\begin{table}[t]
\scalebox{0.7}{
\begin{tabular}{lcccccccc}
\toprule
\multirow{2}{*}{} & \multicolumn{3}{c}{\textbf{Train}}& & \multicolumn{3}{c}{\textbf{Test}} \\ \cmidrule{2-4} \cmidrule{6-8}
& \#Doc    & \#BD   &  \#Types &  & \#Doc    &  \#BD     &  \#Types       \\ \midrule
\textbf{Inter}                             & 18,966     & 529,378  & 24/12   & & 4,652      & 167,087      & 20/10     \\ 
\textbf{Intra}                      & 20,187     & 575,687  & 24/12   & & 4,554      & 163,647      & 20/10      \\ 
 \bottomrule
\end{tabular}}
\label{table_data1}
\caption{
Statistics of the inter- and intra-learning settings in the Few-SEAB dataset, including the numbers of documents (Doc), bounding boxes (BD), and types of entities/relations (Types).
}
\end{table} 

Evaluating the model's transfer learning capacity is crucial for few-shot learning. To conduct the experiment on transfer learning, following \cite{ding-2021-fewnerd}, we reorganize the SEAB dataset and construct two novel datasets, Few-SEAB (Inter) and Few-SEAB (Intra), based on it by adopting different splitting strategies. We divide the entire set of entities into 4 coarse-grained disjoint subsets, e.g., ``\texttt{Shipper and Consignee}'', ``\texttt{Goods Information}'', ``\texttt{Shipping Information}'', and ``\texttt{Address+Numbers}''. 
In other words, we construct the Few-SEAB (Intra) dataset $\langle \hat{\mathcal{D}}_{\rm{train}}$,$\hat{\mathcal{D}}_{\rm{test}} \rangle $ according to the coarse-grained
types with the principle that the entities in different sets belong to different coarse-grained types. For example, ``\texttt{Shipper and Consignee}'', ``\texttt{Goods Information}'' only appear in  $ \hat{\mathcal{D}}_{\rm{train}}$, and ``\texttt{Shipping Information}'', ``\texttt{Address+Numbers}'' only appear in $\hat{\mathcal{D}}_{\rm{test}}$. This characteristic ensures that the training and test sets have minimal overlap in terms of shared knowledge. In contrast, while the fine-grained entity types in the Few-SEAB (Inter) dataset are mutually disjoint
in $ \hat{\mathcal{D}}_{\rm{train}}$ and $\hat{\mathcal{D}}_{\rm{test}}$, the coarse-grained types are shared, which means both $ \hat{\mathcal{D}}_{\rm{train}}$ and $\hat{\mathcal{D}}_{\rm{test}}$ should contains all four fine-grained types ``\texttt{Shipper and Consignee}'', ``\texttt{Goods Information}'', ``\texttt{Shipping Information}'', and ``\texttt{Address+Numbers}''. Table~\ref{inter-intra} gives the comparison of the results under the Intra and Inter transfer learning settings.

\begin{figure}[t]
\centering
\includegraphics[width=\linewidth]{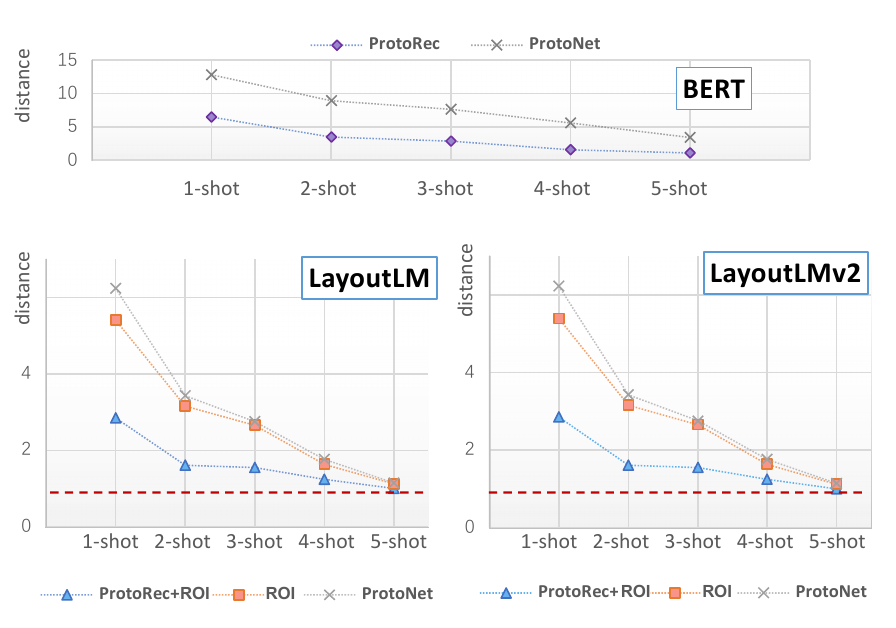}
\caption{Average distance from estimated prototype to class centroid for $K$-shot instances
Distance from the estimated prototype of the $K$-shot instance to the real centroid of the class. We define the distance between the prototype estimated from 5-shots and the actual centroid of the class as 1 and use this as a reference. We report the average distance for all new classes.}
\label{fig:relative_distance}
\end{figure}
\subsection{Effects of Prototypical Rectification}
In the testing phase, the model needs to refer to the mean feature of K-shot examples as the class prototype. As shown in Figure~\ref{fig:relative_distance}, our estimated class prototypes are more robust and accurate than the baseline. We get real class centers for all entity types via supervised tasks. We use all few-shot learning models to compute the relative distance from each model-generated prototype to the real center. The distances to real class centers do not increase too much as the shot decreases because our method can fully leverage base classes’ distributions to estimate novel classes’ distributions. The prototypes sampled from distributions are robust to the variance of support examples. While the baseline is sensitive to the number of support examples.


\section{Related Work}
\subsection{VRD Understanding} 
Researchers have been actively working on addressing the task of understanding VRDs by incorporating multiple modalities, including text, layout, and visual features. In recent years, a variety of approaches have emerged in this field. Grid-based methods, including Chargrid \cite{katti-2018-chargrid}, BertGrid \cite{denk-2019-bertgrid}, and ViBERTgrid \cite{lin-2021-ViBERTgrid}, utilize 2D feature maps for document representation. Graph neural network (GNN)-based methods, such as GraphIE \cite{liu-2019-graph}, MatchVIE \cite{MatchVIE-2021-tang}, and DualVIE \cite{zhang-etal-2022-dualvie}, FormNet \cite{lee-2022-formnet}, FormNetv2\cite{lee-etal-2023-formnetv2} model associations among text segments using graph structures. By leveraging the connectivity between segments, these models can better understand relationships within a document. Transformer-based methods such as LayoutLM \cite{xu-2020-layoutlm} incorporate two-dimensional relative position information based on BERT’s architecture \cite{bert-2018-Devlin}. This enables the model to perceive the positions of text segments within a document. Later works like LayoutLMv2 \cite{xu-2021-layoutlmv2}, StrucText \cite{structtext-2021-li}, StrucTextv2 \cite{structextv2-2023-yu}, LiLT \cite{wang-2022-lilt}, and LayoutLMv3 \cite{layoutlmv3-2022-huang} further integrate visual channel input into a unified multimodal Transformer framework, leading to better alignments and representations across different modalities. These approaches have demonstrated promising progress in improving the performance of understanding VRDs. While \newcite{wang-etal-2021-layoutreader} and \newcite{wang-etal-2023-doctrack} study the effect of reading order on the task of towards human-like machine reading, they did not discuss the local order involving in key-value relationships in the document layout. 

\subsection{Few-shot Relational Learning}
Most recent works on few-shot relational learning in NLP focus on relation extraction or classification using only the text modality \cite{han-2018-fewrel,soares-2019-matching}. These works typically follow the $N$-way $K$-shot setting \cite{snell-2017-protonet}, where a relation instance needs to be assigned to one of $N$ classes based on only $K$ examples per class. Evaluation is done on few-shot relation extraction benchmarks such as FewRel \cite{han-2018-fewrel}, FewRel2.0 \cite{han-2018-fewrel}, and Few-Shot TACRED \cite{sabo-etal-2021-revisiting}. 
While some models have achieved performance surpassing human performance on these tasks, it has been argued by \newcite{brody-etal-2021-towards} and \newcite{sabo-etal-2021-revisiting} that existing benchmarks are far from real-world applications. They emphasize that more challenges remain unsolved in this direction, including reducing reliance on entity type information and focusing more on relations for realistic scenarios. Another challenge is transitioning to document-level relational learning \cite{Popovic2022FewShotDR}, which requires different architectures compared to sentence-level approaches.

Given that the goal of few-shot relation learning is to quickly adapt to unseen relation classes with limited samples by leveraging training on known relation classes, metric learning serves as the primary paradigm for this task. Prominent methods include Prototypical network \cite{snell-2017-protonet,garcia-2018-fewshot} or its variants like prototype rectification \cite{liu-2022-learn}. These methods learn a prototype for each class and classify items based on their similarities to the prototypes. Additionally, efforts have been made to develop frameworks for document-level relational representation learning in visually-rich documents \cite{li-2022-VRD} and enhance prototypes using relation information through prototype rectification modules \cite{liu-2022-learn}, which have also shown promising results. 

\section{Conclusion}
In this paper, we have addressed the research topic of few-shot relational learning in visually-rich documents. Given the limited availability of datasets in this domain, we have reorganized existing supervised benchmark datasets and developed a sampling algorithm specifically tailored for few-shot learning settings. Inspired by human-like cognition, we also propose a novel variational approach to incorporate 2D-spatial priors and relation-agnostic features to improve the model's performance on the few-shot relational learning task. The 2D-spatial priors models the ROI window which guides the model’s attention towards relevant regions for a given relation within the document image. Additionally, we have introduced a prototypical rectification mechanism to enhance its ability to generalize and adapt to new instances despite limited training data. Through extensive experiments conducted on our newly created datasets, we have demonstrated the effectiveness of both our ROI-aware and prototypical rectification techniques in improving performance on few-shot relational learning tasks for VRDs. These advancements significantly contribute to the progress of research in few-shot relational learning for VRDs and pave the way for further exploration in this direction.

\section{Acknowledgements}
This work was supported by National Natural Science Foundation of China (Young Program: 62306173, General Program: 62176153), JSPS KAKENHIProgram (JP23H03454), Shanghai Sailing Program (21YF1413900), Shanghai Pujiang Program (21PJ1406800), Shanghai Municipal Science and Technology Major Project (2021SHZDZX0102), the Alibaba-AIR Program (22088682), and the Tencent AI Lab Fund (RBFR2023012).

\nocite{*}
\section{Bibliographical References}\label{sec:reference}

\bibliographystyle{lrec-coling2024-natbib}
\bibliography{lrec-coling2024-example}

\label{lr:ref}
\bibliographystylelanguageresource{lrec-coling2024-natbib}
\bibliographylanguageresource{languageresource}
\section{Appendix}
\subsection{Relation Types}
\begin{figure}[htbp]
\centering
\includegraphics[width=\linewidth]{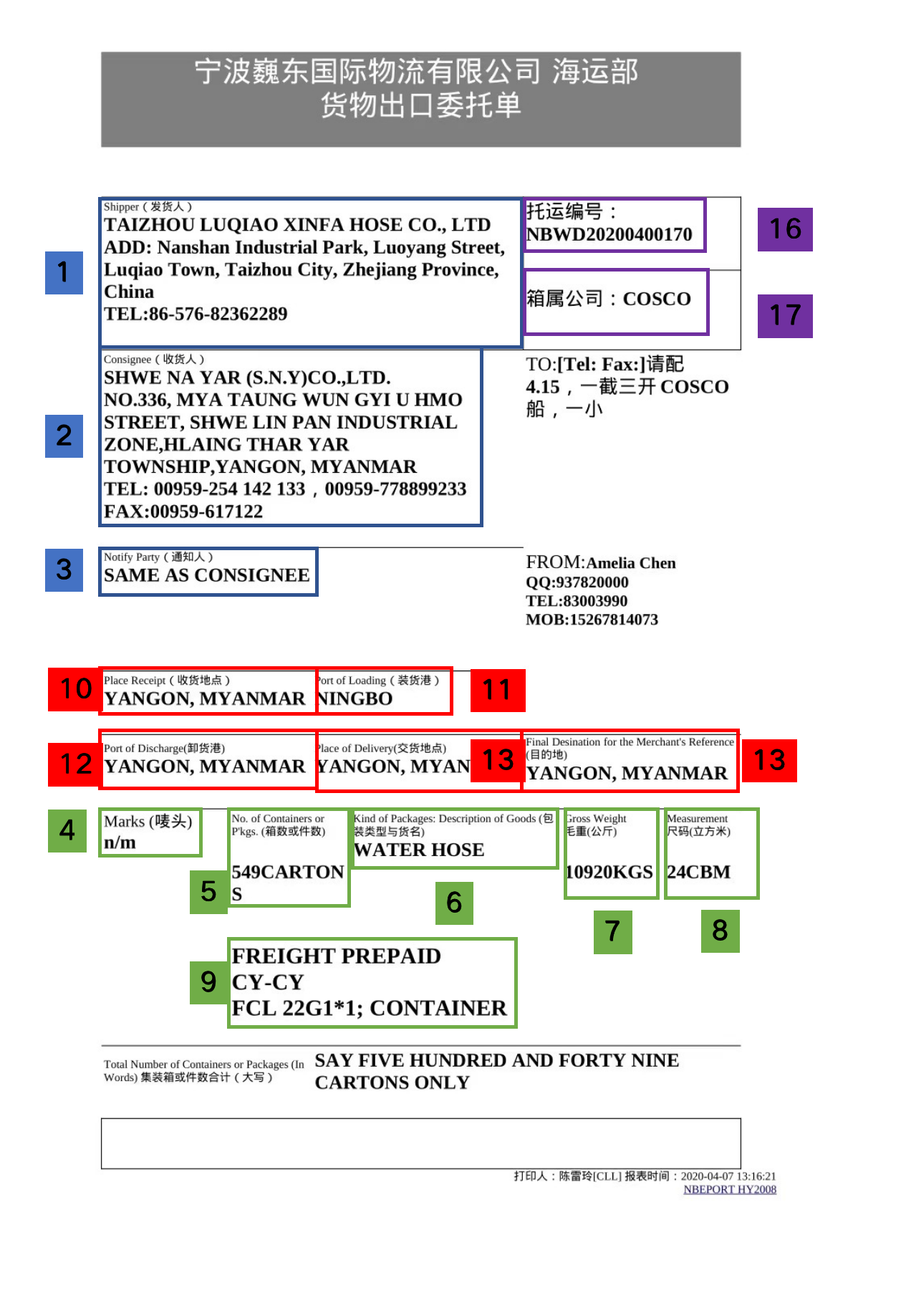}
\includegraphics[width=\linewidth]{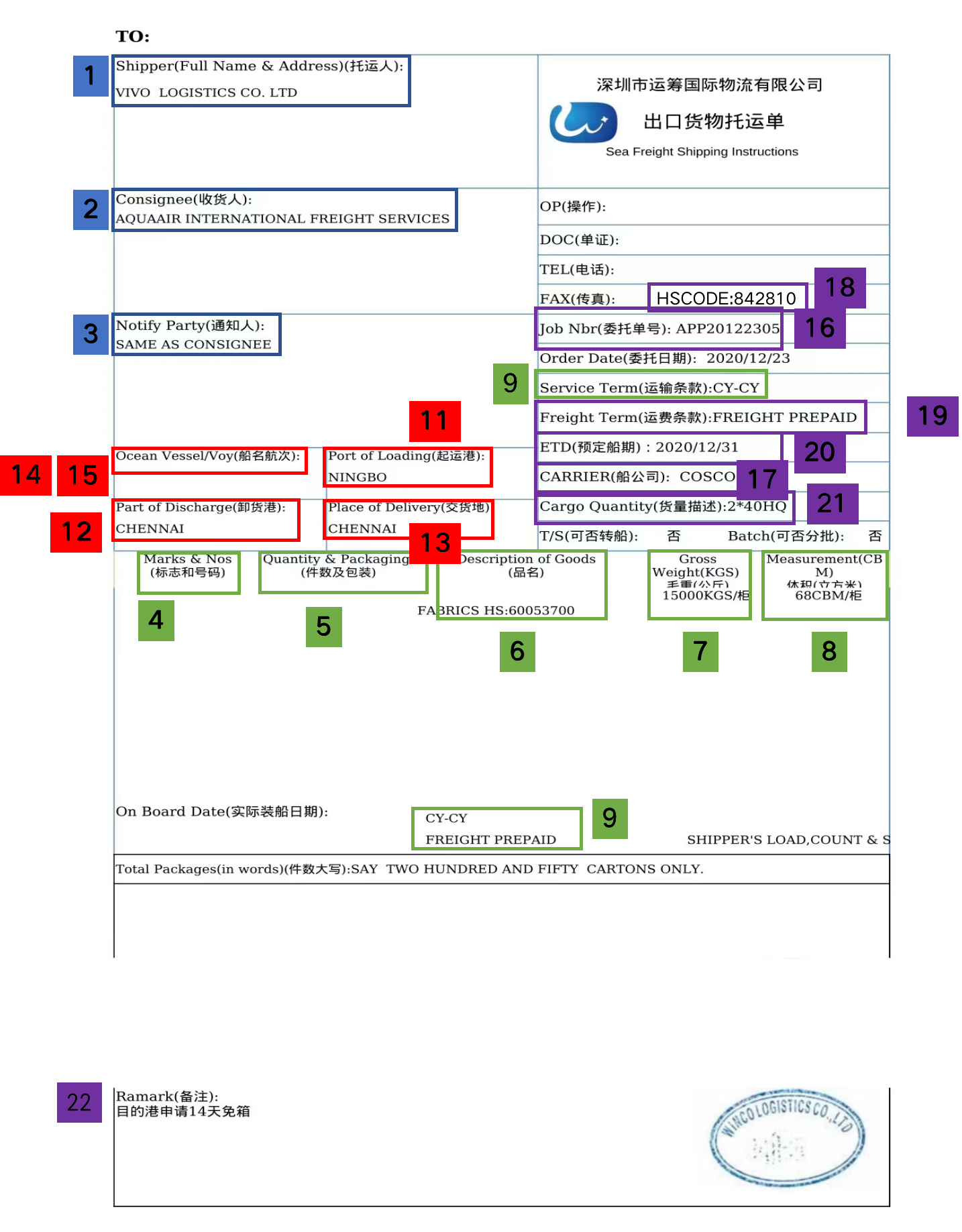}
\caption{Realistic examples with annotations from  the SEAB dataset. These examples show all entity and relation types.}
\label{fig:example}
\end{figure}

As shown in Table~\ref{similarity1}, SEAB is manually annotated with 4 coarse-grained and 22 fine-grained relation types, and we show all the types in Figure~\ref{fig:example}. Note that 
the partition of relation types is shown in the Table~\ref{table_partition}. Different colors is assigned according to their specific coarse-grained types.
\begin{table}[h]
\begin{center}

\scalebox{0.65}{
\begin{tabular}{lcc}
\toprule
{\textbf{}} & {\textbf{\# Train}}& {\textbf{\# Test}} \\ \midrule
SEAB                             & ALL     & ALL \\ \midrule
Inter                          & {\color{blue}1},{\color{green}5},{\color{green}7},{\color{green}8},{\color{green}9},{\color{red}10},{\color{red}11},{\color{red}12},{\color{red}13},{\color{red}14},{\color{red}15},{\color{purple}18}    & {\color{blue}2},{\color{blue}3},{\color{green}4},{\color{green}6},{\color{purple}16},{\color{purple}17},{\color{purple}19},{\color{purple}20},{\color{purple}21},{\color{purple}22}\\ \midrule
Intra                             & {\color{blue}1},{\color{blue}2},{\color{blue}3},{\color{green}9},{\color{red}10},{\color{red}11},{\color{red}12},{\color{red}13},{\color{red}14},{\color{red}15},{\color{red}16},{\color{purple}19} & {\color{green}4},{\color{green}5},{\color{green}6},{\color{green}7},{\color{green}8},{\color{purple}17}, {\color{purple}18},{\color{purple}20},{\color{purple}21},{\color{purple}22}    \\ \bottomrule
\end{tabular}
}
\caption{Partition of relation types in SEAB dataset. Each number corresponds to a relation type. Specifically, 1-22: \texttt{Shipper}, \texttt{Consignee}, \texttt{Notify Party}, \texttt{Marker}, \texttt{Number of Packages}, \texttt{Good Description}, \texttt{Gross Weight}, \texttt{Measurement}, \texttt{Shipping Terms}, \texttt{Place of Receipt}, \texttt{Port of Loading}, \texttt{Port of Discharger}, \texttt{Place of Delivery}, \texttt{Vessel Name}, \texttt{Voyage no}, \texttt{Consignment Code}, \texttt{Shipping Company}, \texttt{HSCODE}, \texttt{Freight Terms}, \texttt{Pre-Assignment}, \texttt{Case Size}, \texttt{Remarks}.
}
\label{table_partition}
 \end{center}
\end{table}

\end{document}